# Development and Preliminary Evaluation of a Domain-Specific Large Language Model for Tuberculosis Care in South Africa

Thokozile Khosa[1], Olawande Daramola[2*]

[1] Department of Computer Science, University of Pretoria, South Africa
[2] Department of Informatics, University of Pretoria, South Africa

`u16073330@tuks.co.za, wande.daramola@up.ac.za`

**Abstract.** Tuberculosis (TB) is one of the world's deadliest infectious diseases, and in South Africa, it contributes a significant burden to the country's health care system. This paper presents an experimental study on the development of a domain-specific Large Language Model (DS-LLM) for TB care that can help to alleviate the burden on patients and healthcare providers. To achieve this, a literature review was conducted to gain an understanding of the current LLM development strategies, specifically within the medical domain. Thereafter, data collection from South African TB guidelines, selected TB literature, and existing benchmark medical datasets was conducted. We performed LLM fine-tuning by using the Quantised Low-Rank Adaptation (QLoRA) algorithm on a medical LLM (BioMistral-7B), and also implemented Retrieval-Augmented Generation using GraphRAG. The developed DS-LLM was evaluated against the base BioMistral-7B model and a general-purpose LLM using a mix of automated metrics and quantitative ratings. The results show that the DS-LLM had better performance compared to the base model in terms of its contextual alignment (lexical, semantic, and knowledge) for TB care in South Africa.

## 1      Introduction

Tuberculosis (TB) is still one of the world's deadliest infectious diseases. Although preventable and curable, the efforts taken to diagnose and treat tuberculosis have not been sufficient to meet the 2030 goals of the World Health Organisation (WHO). This results in the continued transmission of TB, loss of life, and a burden that disproportionately affects low- and middle-income countries. One such country is South Africa, where the incidence of TB is estimated by WHO to be 468 per 100,000 of the population in 2022 [1]. The drivers for such high rates of TB infections include poverty, extreme income inequality, along with other bio-social risk factors, namely HIV co-infection, alcohol abuse, smoking, and diabetes [2].

Worldwide, TB affects 10 million people and is the cause of death of 1.5 million people each year, and South Africa is one of eight countries that contains half of all people with TB [3]. Each year, TB is the cause of thousands of deaths in South Africa and is the leading cause of death from a single infectious agent, only followed by HIV/AIDS [4]. These disease statistics highlight the high burden of TB on the South African health care system and research facilities.

The high prevalence of TB within the Southern African region makes it an important problem that must be tackled holistically. TB affects patients, who rarely complete their TB treatment due to poor continuity of care, along with those close to them [8], [9]. In addition, it affects clinicians, as they must navigate knowledge gaps within the TB domain while operating in an under-resourced healthcare system [2]. Thus, the application of multi-sectoral methods that can help to alleviate this burden on patients and healthcare providers is necessary.

The application of Artificial Intelligence (AI) in healthcare has the potential to revolutionise patient care and disease management by automating tasks, optimising processes, minimising the need for manual intervention, and simplifying routine operations for all stakeholders [5], [6]. Large Language Models (LLMs) are machine learning models that are trained on large textual datasets to generate text. They can be used to assist medical research by documenting and summarising medical literature, and can aid clinical decision support by diagnosing complex cases based on symptoms and medical history [7]. Also, LLMs could automate the handling of patient inquiries, such as advising on medication schedules, which is especially useful in areas that have had limited access to healthcare workers [7].

However, unlike a general-purpose LLM, a domain-specific LLM (DS-LLM) for TB care and management will provide an easily accessible knowledge retrieval system that can answer clinical questions, aid in clinical decision-making by providing researched TB care methods, extract valuable information from clinical notes or publications to aid patient care or studies, and improve access to appropriate TB guidelines to clinicians and patients in South Africa. This will help improve the accurate dissemination of knowledge to patients for better continuous care, help researchers stay up to date with relevant literature to allow them to better address the gaps in South African TB care, and reduce the burden on clinicians. So far, a DS-LLM for TB care in South Africa does not exist. Thus, as a contribution, this study presents an empirical investigation of the feasibility of DS-LLM of TB care in South Africa. The paper reports how fine-tuning an LLM for TB using data sources that include research papers and national health guidelines provided by the South African Department of Health is a promising application of AI to improve the experience of TB healthcare in South Africa.

The rest of this paper is described as follows. Section 2 presents the background and a review of related work. Section 3 presents the methodology adopted for this study, while the experimental results are presented in Section 4. The results are discussed in Section 5, while the conclusion and the future work are presented in Section 6.

## 2 Background and Related Work

This presents relevant theoretical background for this study and a review of related work.

### 2.1 Domain-Specific Large Language Models

Large Language Models (LLMs) are generative AI systems that can generate original content, such as images, music, or text, instead of simply analysing patterns found in the data. LLMs that focus on text generation perform text-related tasks such as editing articles, learning languages, generating code, and more [10]. LLMs are deep learning models trained on substantial amounts of text data, enabling them to understand and generate natural human language. Most LLMs utilise transformer-based architecture, which has dramatically improved their natural language processing capabilities [11].

Fine-tuning is a technique used in Natural Language Processing (NLP) to improve the performance of a pre-trained language model on specific tasks, including domain-specific applications [12]. Transfer learning is a key technique for fine-tuning LLMs because it enables the reuse of a pre-trained model for a specific task by retraining the model on a smaller, task-specific dataset to update its parameters. During the fine-tuning process, certain neural network layers of the pre-trained model are frozen while the remaining layers are updated for the specific downstream task. This helps in achieving the required domain-specific model accuracy while limiting the computational cost of training a model from scratch [12].

Several types of fine-tuning techniques can be implemented depending on the desired application. One is unsupervised learning, which uses unlabelled textual data from the target domain for training. Another is supervised learning, which instead uses labelled textual data from the target domain [13]. This approach is more cumbersome as substantial labelled data is required, but it is the optimal method for classification tasks. Finally, there is prompt engineering, which uses natural language instructions to the model to create a domain-specific assistant. Prompt engineering requires no additional training for the model and instead relies on constructing high-quality prompts for optimal application. Similarly, in-context learning enables the model to perform new tasks using the additional information from a given prompt. Few-shot, one-shot, and zero-shot prompting are examples of in-context learning methods [13], [14].

Full fine-tuning is performed when all the model parameters are updated to allow a comprehensive adaptation to the new task. An alternative approach is half fine-tuning (HFT) or Parameter-Efficient Fine-Tuning (PEFT) [15], where a small subset of the model's parameters is updated. PEFT approaches, such as adapter tuning, attach an additional trainable layer to the pre-trained model's layers to enable task-specific adaptation. This helps reduce computational costs and overfitting associated with full fine-tuning [14].

Additional considerations that can be made when fine-tuning for a specific task include applying quantisation to the model parameters to reduce the memory requirements and applying gradient accumulation, where gradients are calculated only after multiple mini-batches before updating the model. PEFT methods like LoRA and QLoRA can also be used when fine-tuning using smaller datasets [12].

Results by [16] showed that the fine-tuning process reduced their model's performance in language generation, even though validation loss decreased with training. One method proposed to improve the accuracy of the generated answers from LLMs is by using Retrieval Augmented Generation (RAG). RAG extends the model's generation capabilities on tasks that require high factuality by introducing contextual domain-specific information from external datasets [16].

## 2.2 Related Work

The study conducted by [17] explored developing a domain-specific large language model, focusing on prostate cancer. Here, clinical notes from patients treated for prostate cancer were collected, including radiology and pathology reports, and used as the training dataset. A pre-training strategy was employed. Recall@K was used to evaluate the masked clinical information, and a curated set of clinical questions and answers (Q&A) from the American Cancer Society guidelines was used to assess accuracy on the domain knowledge. Comparisons between BioGPT, GPT-2, and the developed PCa-LLM showed that the custom model outperformed GPT-2 and BioGPT at each K value for masked clinical information retrieval and in the user study on the Q&A.

Another study by [18] explored fine-tuning different open-source LLMs for radiation oncology tasks and assessed the developed models on specific tasks. It made use of patient cases as training data and used the PEFT technique, Low-Rank Approximations (LoRA) on two open source LLMs, LLaMA2-7B and Mistral-7B. The fine-tuned models were evaluated using ROUGE-1 and accuracy metrics. The results highlighted that the fine-tuned model outperformed vanilla models across all three tasks, demonstrating that fine-tuning LLMs with domain knowledge can significantly improve their performance in healthcare tasks.

Similarly, [19] used fine-tuning to develop a medical domain-specific LLM, but to address answering ophthalmology-related patient queries. Further highlighting the advantages of fine-tuning when building LLMs for medical subspecialties.

With the rise in RAG and recognition of its advantages over fine-tuning, other studies started making use of RAG to improve medical LLM implementations.

The study by [20] proposed integrating a specialised Large Language Model into a digital adherence technology (DAT) to improve communication and treatment outcomes for TB patients. This study aimed to develop an LLM-powered TB treatment support tool using real-world Argentinian data and in-context learning techniques. Additionally, it sought to evaluate the developed model on linguistic appropriateness, empathy, medical accuracy, and privacy.

TB guidelines, medication suggestions, previous TB trial messages, and manually crafted dialogue samples were used as data sources for a Retrieval Augmented Generation (RAG) system. The domain adaptation was explored using prompt engineering techniques with a GPT model. Human evaluations were used to assess empathy, medical accuracy, and linguistic accuracy in the generated responses.

Another study explored improving RAG implementations specifically for medical applications [21]. The authors proposed a novel graph-based RAG framework designed for the medical domain called MedGraphRAG. This was because general RAG designs struggled with getting insights across documents, while GraphRAG is unable to verify the responses. The goal was a custom construction that could enhance LLM capabilities for generating evidence-based medical responses.

Different RAG architectures were compared to the novel RAG implementation using human evaluators. The comparison included standard RAG methods implemented using LangChain [22] and GraphRAG [23] implemented using Microsoft Azure. The findings showed that MedGraphRAG enhanced LLM performance on health fact-checking and medical Q&A benchmarks compared to baseline LLMs (without retrieval) and other RAG methods. Additionally, this approach outperformed intensively fine-tuned medical LLMs.

This study is uniquely positioned in that it aims to focus on TB within the South African context and combine fine-tuning techniques and RAG techniques to improve the domain alignment of the large language model. Additionally, adapting for general TB question-answering and medical summarisation tasks.

## 3 Methodology

We adopted an experimental research design for this study (see Figure 1). An experimental research approach is appropriate because it can help to determine the most appropriate data collection method, data processing techniques, algorithm design, and evaluation strategies. The different stages of the experimental process adopted for this study are presented next.

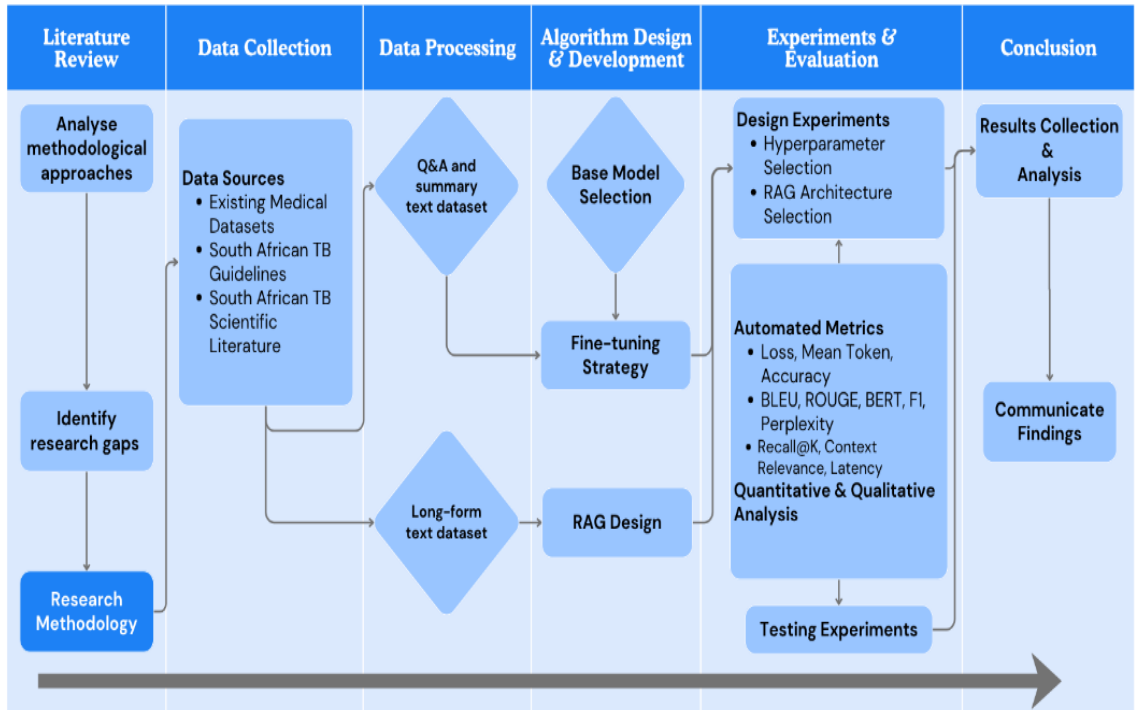

**Fig. 1.** Experimental Research Design Framework for the Study

### 3.1 Data Collection

High-quality textual medical data is a foundational component for medical large language models. For this reason, the variety and scope of specialised medical datasets have increased to aid research and development. This data can be in the form of conventional text data, which enables models to understand and generate human language or can be multimodal, like those containing images and time series data. For this study, open-source public medical datasets, clinical guidelines and scientific literature from PubMed and PubMed Central (PMC) were sourced to be used as medical textual data.

**TB Medical Guidelines**
A range of South African clinical guidelines, management guidelines, fact sheets, and healthcare providers' training manuals focused on TB were collected from various websites, including those of the South African Department of Health, Knowledge Hub Health, and the National Institute for Communicable Diseases (NICD), among others. These were all collected in PDF format to undergo data extraction and processing.

**TB Scientific Literature**
PubMed [24] was used to obtain scientific abstracts on TB publications that focus on South Africa, and PMC [25] was used to extract the full text of TB-related medical literature published by South African institutions in 2025. This was achieved by using the E-utilities API to extract the abstracts and full text in XML format. The year of articles selected was chosen after the model selection, to ensure no redundant literature was used for this fine-tuning process.

**Benchmark Q&A Datasets**

Existing high-quality medical datasets were leveraged as they provided medical textual data already in Q&A format and verified in other studies. Though not focused on the South African context, this dataset can provide medical question-answering capabilities that are required from the model. This makes it a good complementary dataset. The extracted benchmark datasets were filtered for TB keywords to create a subset dataset with relevant question-answer pairs.

AfriMedQA [26], MedQA [27], PubMedQA [28], MedMCQA [29], and MMLU-Medical [30] were all aggregated and used in this study. These five benchmark datasets are the most prominent benchmarks used in literature when training or evaluating medical knowledge in LLMs. Additional analysis by [31] revealed that the representation of diseases that affect Africans at a disproportionate rate is low in these datasets. For tuberculosis specifically, AfriMedQA had the highest representation for TB from these open-source medical datasets [31].

### 3.2 Data Preprocessing

The collected data were re-formatted into an instruction fine-tuning structure to enable question-answering, instruction generation and summarisation tasks.

The clinical guidelines' textual data were formatted into an instruction fine-tuning dataset using a mix of keywords and headings as instruction methods [12]. This was a lightweight text reformatting strategy that enables the model to expand its capabilities beyond next token generation.

Considering that the clinical guidelines are a crucial textual dataset for this research, methods for extracting pure question-answer pairs from the dataset were also explored. The data augmentation and prompting strategies outlined by [32] were implemented with a TB-centred modification to the prompting strategy. This was used to create long-form question-answer pairs for the training dataset [32].

The TB literature (abstracts) was structured for summarisation tasks by using the title and abstract text as an input/output pair in the instruction prompt. The full-text TB literature was subdivided per journal/article section and kept as raw text chunks. Using a similar prompt as that used in the clinical guidelines and randomly sampled text chunks, a PubMed article's long-form question-answer pair was generated to assess the RAG implementation.

The question-answer benchmark datasets were used as is, as they are already in question-answer structure.

The complete ETL process and architecture for the text extraction, data cleaning, data validation, structural transformation, and data splitting are summarised in Figure 2. Here, the final dataset was subdivided into the fine-tuning dataset and the RAG dataset.

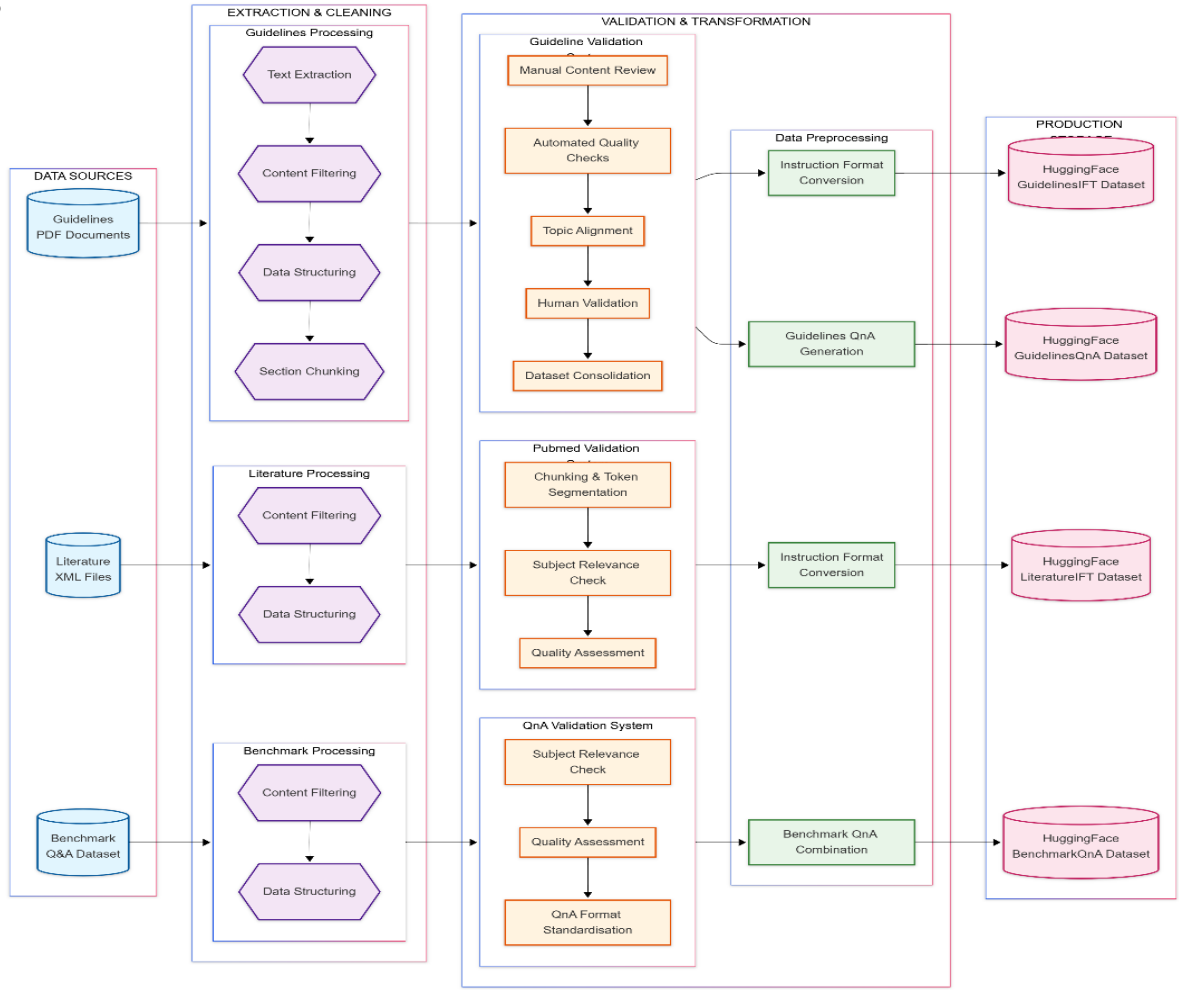

**Fig. 2.** Data Processing Pipeline

### 3.3 Algorithm Design and Development

**Base model selection**

Various medical LLMs were analysed as potential models to use in this study. These models were identified from various medical LLMs leaderboards [33], [34], [35] that compare the performance of various LLMs (medical and general) to prominent medical benchmarks. Additionally, [36] conducted a survey outlining existing medical-domain LLMs and their developmental progress. From these sources, well-performing, open-source models with available literature on the models' development process were identified. Another consideration was the availability of small parameter-size model variants (ten billion parameters and less due to RAM limitations).

The BioMistral 7B-DARE model [37] was selected as the base model for this study due to its better ranking on the Hugging Face medical LLM leaderboard and its performance in the literature in comparison to the other medical LLMs. The BioMistral model range also has multiple size options to accommodate different hardware limitations (i.e. can test code with smaller models to minimise GPU wastage).

### 3.4 Fine-Tuning Strategy

From the available fine-tuning techniques, QLoRA (Quantised Low-Rank Adaptation) and instruction fine-tuning present the most advantages for our use case. QLoRA, which is a type of Parameter-Efficient Fine-Tuning (PEFT) technique, has smaller hardware requirements (GPU memory) without compromising on model performance, and it is prominently documented in its application within the literature. Additionally, due to the smaller size of the training dataset, full fine-tuning is not a viable option [12]. Instruction fine-tuning will also aid in aligning the model to the domain-specific tasks. QLoRA, when implemented on a pre-trained medical LLM, can enable data efficiency, since a smaller fine-tuning dataset can be used to expand on the current medical knowledge of the baseline model.

### 3.5 GraphRAG Architecture

The functionality of the model was to be expanded upon by implementing a text-based scientific GraphRAG system [38] to enhance factuality when posed with clinical or research-based TB queries. GraphRAG performs entity extraction using biomedical name entity recognition (NER) models and graph linking using an embedding approach.

A hybrid heuristic-based retriever was implemented to identify and retrieve the relevant information from the graph-structured data. This combined a similarity-based retriever that uses BM25 [39] to perform lexical searching (exact word searching using term frequency and inverse document frequency) and dense vector similarity searching using the FAISS (Facebook AI Similarity Search) embedding to find semantically similar documents, a knowledge graph-based semantic retriever, and graph-based entity-expansion to find neighbouring relevant documents [38], [40].

### 3.6 Hyperparameter Optimisation

We used LoRA with 4-bit quantisation (QLoRA) to fine-tune the 7B Mistral LLM. Hyperparameters such as LoRA Rank, Alpha, Dropout, Batch Size, Learning Rate, Warm-up Ratio and Weight Decay were optimised using the Optuna framework [41], [42].

### 3.7 Model Evaluation

The evaluation scenarios were completed on the fine-tuned model and the implementation of RAG using the test split in the dataset. These were all under zero-shot conditions to enable the automation on the full dataset.

Automated metrics, in addition to quantitative evaluation scenarios, are required for a holistic evaluation of this type of model. The quantitative approach was used to assess the accuracy and factuality dimensions of the generated text.

To standardise this process, a rating rubric was developed to assist in the evaluation of the generated answers from the model. The rubric was based on similarly developed rubrics as seen in literature and customised for this model's stage of development [20], [43], [44].

## 4 Results

This section presents the results from the LLM fine-tuning and the RAG implementation.

### 4.1 Fine-Tuned Model Results

The fine-tuned model (BioMistral-7B-TB-Optimised) and base model (BioMistral-7B-DARE) were evaluated on the test dataset (unseen) at inference, with the following metrics automatically calculated per generated answer against the reference/ benchmark answer. All these results use a zero-shot prompting approach.

Metrics such as ROUGE-L (range 0-1) and F1 (range 0-1) require strict generated text matching to get high matching scores. They do, however, still allow a fair comparison to be conducted between the base model and fine-tuned model to measure phrasing/answering improvements. The metric BERT-F1 (range 0-1) is better suited to check the meaning alignment in the generated answers, regardless of re-phrasing. This score was recomputed using a biomedical embedding model, BioBERT, for better accuracy on the subject matter [105].

Perplexity_pred (range 1-infinity) is suited to check the confidence in the model's predicted text sequence. With a score closer to one, meaning more confidence. The mean results per dataset are summarised in Table 1.

**Table 1.** Models' Testing Performances (ROUGE-L, F1, BERT-F1, Perplexity)

| Dataset | ROUGE-L | | F1 | | BERT-F1 | | $PPL_{pred}$ | |
|---|---|---|---|---|---|---|---|---|
| | DARE | TB | DARE | TB | DARE | TB | DARE | TB |
| Guidelines Q&A | 0.2420 | 0.2980 | 0.3790 | 0.4720 | 0.9625 | 0.9750 | 2.604 | 2.365 |
| Guidelines IFT | 0.1592 | 0.1600 | 0.2407 | 0.3180 | 0.9296 | 0.9201 | 2.370 | 2.057 |
| Benchmark Q&A | 0.05874 | 0.1700 | 0.0617 | 0.2960 | 0.6505 | 0.9052 | 82.48 | 2.330 |
| PubMed Abstract | 0.1641 | 0.1740 | 0.2938 | 0.3780 | 0.9548 | 0.9674 | 2.530 | 2.445 |

### 4.2 Retrieval-Augmented Generation Testing Results

The fine-tuned model was similarly connected to the hybrid GraphRAG schema and fed the RAG testing dataset to assess the performance of the model with retrieved context. To get a second comparative view of the RAG performance isolated from the fine-tuned model, the hybrid GraphRAG schema was connected to the GPT-4o-mini model using OpenAI's API. This was also evaluated on the same testing dataset.

The GPT-4o-mini model was chosen because, in the paper by [37] that outlines the BioMistral range design, the base model (BioMistral-DARE) was also evaluated against the GPT-3.5 Turbo model. This was the only model in the papers' evaluation set that outperformed the base model overall [37]. The 4o-mini was used in this case since it is a comparable and more recent variant with lower API usage costs.
The mean results per metric are summarised in Table 2.

**Table 2.** Retrieval and Efficiency Metrics (GraphRAG Evaluation)

| Model | Recall@K | Context Relevance | Entities Used | Latency (s/it) |
|---|---|---|---|---|
| **PMC** | | | | |
| BioMistral-7B-TB | 0.9538 | 0.9962 | 25.53 | 45.83 |
| GPT-4o-mini | 0.9538 | 0.9962 | 25.53 | 29.51 |
| **Guidelines** | | | | |
| BioMistral-7B-TB | 0.7400 | 0.9950 | 17.20 | 52.17 |
| GPT-4o-mini | 0.7400 | 0.9950 | 17.20 | 29.18 |

### 4.3 Accuracy and Factuality Ratings

Accuracy and Factuality results were collected for each generated answer and compared to the reference answer. This was completed only for the question-answer task datasets, as these can be quantified for accuracy.

Each result was assessed using an LLM-as-a-Judge Prompt structure outlined in [45], [46], adjusted for accuracy and factuality. This provides a better assessment of each question-answer pair. The combined rating results were rescaled into a percentage for easier comparison. This is summarised in Table 3.

**Table 3.** Accuracy and Factuality Metrics Across Datasets

| Model | Benchmark | | Guidelines | | PubMed | | Average | |
|---|---|---|---|---|---|---|---|---|
| | Acc. | Fact. | Acc. | Fact. | Acc. | Fact. | Acc. | Fact. |
| BioMistral-7B-DARE | 50.44 | 52.02 | 63.40 | 72.60 | - | - | 56.92 | 62.31 |
| BioMistral-7B-TB | 54.82 | 60.79 | 71.20 | 77.20 | - | - | 63.01 | 69.00 |
| BioMistral-7B-TB + GraphRAG | - | - | 71.40 | 79.40 | 68.00 | 76.00 | 69.70 | 77.70 |
| GPT-4o-mini+ GraphRAG | - | - | 68.00 | 78.40 | 63.08 | 72.31 | 65.54 | 75.36 |

## 5 Discussion

The evaluation results of our study are discussed in terms of the automated performance and quantitative performance.

The testing sequence revealed improvements in the automated scores such as ROUGE-L, F1, BERT-F1, and perplexity. This was for most of the test datasets, which shows a better lexical and semantic alignment with the TB-specific medical text.

The performance improvements were mostly evident in the benchmark dataset, highlighting the adaptation to TB question-answering tasks. The other dataset improvements, related to the summarisation tasks and the guidelines question-answering, were more modest in comparison. This is potentially due to a limitation in the data diversity of the curated South African dataset.

The construction and addition of a knowledge graph RAG proved beneficial in extracting relevant context and linked entity information from the vector store. The testing sequence highlighted that the addition of context-relevant information aided the general generator (GPT-4o-mini) with lexical and semantic alignment, but not so much the fine-tuned model's alignment, which decreased slightly or stayed the same.

The quantitative evaluation results used a rating system to compare the baseline model, the fine-tuned model, and the integration of GraphRAG. The developed rating system highlighted improvements in the accuracy and factuality of the fine-tuning process. Additionally, it highlighted the improved factuality that the RAG system provides to both the general and fine-tuned models. The combination of the fine-tuned model and the RAG system also indicates better reasoning capabilities over the contextual information due to the model having better alignment with TB-specific lexical and semantic language.

### 5.1 Limitations

Due to limited computing resources and hardware, smaller variants of medical models were explored as the base model. The models that were going to be feasible for use needed to fit within the limitations of a Google

Colab environment. This resource limitation also restricted the training strategies that could be experimented with.

Additionally, a TB specialist was not involved in this process. Instead, the current work focused on the technical feasibility and exploring viable fine-tuning frameworks from the literature and through experimentation for model adaptation. Future work will involve TB domain experts for directions on additional data sources, working with larger datasets, including academic publications covering several years, not just 2025, ensuring alignment with clinical practices and using healthcare workers to assess the generated results of the model.

# 6      Conclusion

This study developed a domain-specific LLM to assist in providing accurate TB information and guidelines to patients and practitioners in the South African context. The study objectives of identifying optimal fine-tuning methodologies from literature, curating a custom South African TB dataset, training a domain-specific LLM and evaluating the model performance were met.

The study explored the use case and benefits of fine-tuning a medical LLM using a curated TB-specific dataset, together with the integration of a hybrid GraphRAG system, in adapting a large language model for Tuberculosis care. The adopted experimental design approach allows for the production of a domain-aligned system with improved TB question-answering and summarisation capabilities. This work provides a practical stepping stone for the development of more accurate region-specific TB-focused LLMs to support the South African healthcare system.

Regarding contributions, theoretically, this study enables a better understanding of how domain adaptation methods, specifically QLoRA, and context knowledge retrieval and integration with GraphRAG, can positively improve the performance of a general medical LLM on TB-specific tasks. In addition, this study adds to the understanding of how different TB data sources and their formatting influence the LLM adaptation to different clinical and research tasks. Methodologically, this study contributes a TB-specific data collection and data processing pipeline, from identifying reliable sources to cleaning the non-digitised textual data and formatting it for both instruction fine-tuning and GraphRAG construction. Additionally, it provides a custom combination of fine-tuning strategies that combine different TB textual data sources, instruction fine-tuning, quantised LoRA fine-tuning and a scientific GraphRAG construction with customised prompting. Practically, this study contributes a custom South African TB dataset for training and evaluation of LLMs. The developed domain-specific LLM will serve as a decision-support tool for TB care practitioners and researchers.

In future work, we will focus on the expansion of the quantity and diversity of the fine-tuning training dataset. We will include South African medical training test questions, as well as datasets validated by TB researchers and experts, for question-answering purposes. The model training process will be further explored by experimenting with larger models. Finally, the inclusion of human evaluators who are medical specialists and general healthcare workers for more rigorous evaluation scenarios.

# References

[1]     *Global Tuberculosis Report 2023*, 1st ed. Geneva: World Health Organization (2023).
[2]     Izrah M: A spotlight on the tuberculosis epidemic in South Africa. National Communication, 15(1): 1290. (2024). doi: 10.1038/s41467-024-45491-w.
[3]     Tuberculosis: Accessed: Oct. 11, 2025. [Online]. Available: https://www.who.int/health-topics/tuberculosis
[4]     Mineral Council South Africa: Tuberculosis in the South African Mining Industry, [Online].Available: https://www.mineralscouncil.org.za/component/jdownloads/?task=download.send&id=769:tuberculosis-in-south-africa&catid=&m=0
[5]     Davenport T, Kalakota R: The Potential for Artificial Intelligence in Healthcare. *Future Healthc. J.*, 6(2): 94–98 (2019). doi: 10.7861/futurehosp.. 6-2-94.

[6]   Panagoulias DP, Sotiropoulos DN, Tsihrintzis GA: SVM-Based Blood Exam Classification for Predicting Defining Factors in Metabolic Syndrome Diagnosis. *Electronics*, 11(6): 857–857 (2022). doi: 10.3390/electronics11060857.

[7]   Anisuzzaman DM, Malins JG, Friedman PA, Attia ZI: Fine-Tuning Large Language Models for Specialized Use Cases. *Mayo Clin. Proc. Digit. Health*, 3(1): 100184 (2025). doi: 10.1016/j.mcpdig.2024.11.005.

[8]   Dudley L, Mukinda F, Dyers R, Marais F, Sissolak D: Mind the gap! Risk factors for poor continuity of care of TB patients discharged from a hospital in the Western Cape, South Africa. *PLOS ONE* 13(1): e0190258 (2018). doi: 10.1371/journal.pone.0190258.

[9]   Kallon II, Colvin CJ, Trafford Z: A qualitative study of patients and healthcare workers' experiences and perceptions to inform a better understanding of gaps in care for pre-discharged tuberculosis patients in Cape Town, South Africa. *BMC Health Serv. Res.* 22(1):128 (2022). doi: 10.1186/s12913-022-07540-2.

[10]  Corchado JM, López S, Garcia R, Chamoso P: Generative artificial intelligence: Fundamentals. ADCAIJ: advances in distributed computing and artificial intelligence journal, 12, e31704-e31704 (2023).

[11]  Minaee S, Mikolov T, Nikzad, N, Chenaghlu M, Socher R, Amatriain X, Gao J. Large language models: A survey (2024). arXiv preprint arXiv:2402.06196.

[12]  VM K, Warrier H, Gupta Y: Fine tuning llm for enterprise: Practical guidelines and recommendations (2024). arXiv preprint arXiv:2404.10779.

[13]  Patil R, Gudivada V: A review of current trends, techniques, and challenges in large language models (llms). Applied Sciences, 14(5): 2074 (2024).

[14]  Parthasarathy VB, Zafar A, Khan A, Shahid A: The ultimate guide to fine-tuning llms from basics to breakthroughs: An exhaustive review of technologies, research, best practices, applied research challenges and opportunities. *arXiv preprint arXiv:2408.13296* (2024).

[15]  Hui T, Zhang Z, Wang S, Xu W, Sun Y, Wu H: Hft: Half fine-tuning for large language models. In *Proceedings of the 63rd Annual Meeting of the Association for Computational Linguistics (Volume 1: Long Papers)*. 12791-12819 (2025).

[16]  Li J, Yuan Y, Zhang Z: Enhancing llm factual accuracy with rag to counter hallucinations: A case study on domain-specific queries in private knowledge-bases. *arXiv preprint arXiv:2403.10446* (2024).

[17]  Tariq A, Luo M, Urooj A, Das A, Jeong, J, Trivedi S, ... , Banerjee I: Domain-specific llm development and evaluation–a case-study for prostate cancer. *medRxiv*, 2024-03 (2024).

[18]  Wang P, Liu Z, Li Y, Holmes J, Shu P, Zhang, L, ... , Liu W: Fine-tuning open-source large language models to improve their performance on radiation oncology tasks: A feasibility study to investigate their potential clinical applications in radiation oncology. *Medical physics*, *52*(7), e17985 (2025).

[19]  Tan TF, Elangovan K, Jin L, Jie Y, Yong L, Lim J, ..., Ting DSW: Fine-tuning large language model (llm) artificial intelligence chatbots in ophthalmology and llm-based evaluation using GPT-4. *arXiv preprint arXiv:2402.10083* (2024).

[20]  Filienko D, Nizar M, Roberti J, Galdamez D, Jakher H, Iribarren S, ..., De Cock M: Transforming Tuberculosis Care: Optimizing Large Language Models For Enhanced Clinician-Patient Communication. *arXiv preprint arXiv:2502.21236* (2025).

[21]  Wu J, Zhu J, Qi Y, Chen J, Xu M, Menolascina F, Grau V: Medical graph rag: Towards safe medical large language model via graph retrieval-augmented generation. *arXiv preprint arXiv:2408.04187* (2024).

[22]  Bratanic T: *Enhancing RAG-based applications accuracy by constructing and leveraging knowledge graphs* (2024) [Online] Accessed: July 07, 2025.. Available: https://blog.langchain.com/enhancing-rag-based-applications-accuracy-by-constructing-and-leveraging-knowledge-graphs/

[23]  Edge, D, Trinh H, Cheng N, Bradley J, Chao A, Mody A, ... , Larson J: From local to global: A graph rag approach to query-focused summarization. *arXiv preprint arXiv:2404.16130* (2024).

[24]  PubMed Data: PubMed. [Online]. Available: https://pubmed.ncbi.nlm.nih.gov/download/

[25]  PubMed Central (PMC): PubMed Central (PMC). [Online]. Available: https://pmc.ncbi.nlm.nih.gov/

[26]  Nimo C, Olatunji T, Owodunni AT, Abdullahi T, Ayodele E, Sanni M, ..., Asiedu MN: AfriMed-QA: a Pan-African, multi-specialty, medical question-answering benchmark dataset. In *Proceedings of the 63rd Annual Meeting of the Association for Computational Linguistics (Volume 1: Long Papers)*,1948-1973 (2025).

[27]  Jin D, Pan E, Oufattole N, Weng WH, Fang H, Szolovits P: What disease does this patient have? a large-scale open domain question answering dataset from medical exams. *Applied Sciences*, *11*(14): 6421 (2021).

[28] Jin Q, Dhingra B, Liu Z, Cohen W, Lu, X: Pubmedqa: A dataset for biomedical research question answering. In *Proceedings of the 2019 conference on empirical methods in natural language processing and the 9th international joint conference on natural language processing (EMNLP-IJCNLP)*, 2567-2577 (2019).

[29] Pal A, Umapathi LK, Sankarasubbu M: Medmcqa: A large-scale multi-subject multi-choice dataset for medical domain question answering. In *Conference on health, inference, and learning.* 248-260 (2022). PMLR. [Online]. Available: https://proceedings.mlr.press/v174/pal22a.html

[30] Hendrycks D, Burns C, Basart S, Zou A, Mazeika M, Song D, Steinhardt J: Measuring massive multitask language understanding. *arXiv preprint arXiv:2009.03300* (2020).

[31] Mutisya F, Gitau S, Syovata C, Oigara D, Matende I, Aden M, ..., Chidede T: Mind the Gap: Evaluating the Representativeness of Quantitative Medical Language Reasoning LLM Benchmarks for African Disease Burdens. *arXiv preprint arXiv:2507.16322* (2025).

[32] Wei L, Ying Z, He M, Chen Y, Yang Q, Hong Y, ..., Chen Y. Diabetica: Adapting Large Language Model to Enhance Multiple Medical Tasks in Diabetes Care and Management. *arXiv preprint arXiv:2409.13191* (2024).

[33] Dilmegani C: Compare 9 Large Language Models in Healthcare (2025). AIMultiple. Accessed: Aug. 31, 2025. [Online]. Available: https://research.aimultiple.com/large-language-models-in-healthcare/

[34] 'Vals AI'. Accessed: Aug. 31, 2025. [Online]. Available: https://www.vals.ai/

[35] 'openlifescienceai (Open Life Science AI)'. Accessed: Aug. 31, 2025. [Online]. Available: https://huggingface.co/openlifescienceai

[36] Zhou H, Liu F, Gu B, Zou X, Huang J, Wu J, ..., Clifton DA: A survey of large language models in medicine: Progress, application, and challenge. *arXiv preprint arXiv:2311.05112*. (2023).

[37] Labrak Y, Bazoge A, Morin E, Gourraud PA, Rouvier M, Dufour R: Biomistral: A collection of open-source pretrained large language models for medical domains. *arXiv preprint arXiv:2402.10373* (2024). doi: 10.48550/arXiv.2402.10373.

[38] Han H, Wang Y, Shomer H, Guo K, Ding, J., Lei, Y., ... & Tang, J: Retrieval-augmented generation with graphs (graphrag). *arXiv preprint arXiv:2501.00309* (2024).

[39] Robertson S, Zaragoza H: The probabilistic relevance framework: BM25 and beyond. *Foundations and Trends® in Information Retrieval*, *3*(4): 333-389 (2009).

[40] Oche AJ, Folashade AG, Ghosal T, Biswas A: A Systematic Review of Key Retrieval-Augmented Generation (RAG) Systems: Progress, Gaps, and Future Directions. *arXiv preprint arXiv:2409.15730* (2024). doi: 10.48550/arXiv.2507.18910.

[41] Tribes C, Benarroch-Lelong S, Lu P, Kobyzev I: Hyperparameter optimization for large language model instruction-tuning. *arXiv preprint arXiv:2312.00949* (2023). doi: 10.48550/arXiv.2312.00949.

[42] Akiba T, Sano S, Yanase T, Ohta T, & Koyama M: Optuna: A next-generation hyperparameter optimization framework. In *Proceedings of the 25th ACM SIGKDD international conference on knowledge discovery & data mining.* 2623-2631 (2019).

[43] Seo J, Choi D, Kim T, Cha WC, Kim M, Yoo H, ..., Choi E: Evaluation framework of large language models in medical documentation: Development and usability study. *Journal of Medical Internet Research*, *26*, e58329 (2024). doi: 10.2196/58329.

[44] Chang Y, Wang X, Wang J, Wu Y, Yang L, Zhu K, ... , Xie X: A survey on evaluation of large language models. *ACM transactions on intelligent systems and technology*, *15*(3):1-45 (2024). doi: 10.1145/3641289.

[45] Kanithi PK, Christophe C, Pimentel MA, Raha T, Saadi N, Javed H, ..., Khan S: Medic: Towards a comprehensive framework for evaluating llms in clinical applications. *arXiv preprint arXiv:2409.07314* (2024). doi: 10.48550/arXiv.2409.07314.

[46] Han H, Ma L, Shomer H, Wang Y, Lei Y, Guo K, ..., Tang J: Rag vs. graphrag: A systematic evaluation and key insights. *arXiv preprint arXiv:2502.11371* (2025). doi: 10.48550/arXiv.2502.11371.